\documentclass{article}

\usepackage{authblk}
\usepackage{amsmath}

\usepackage{arxiv}

\usepackage[utf8]{inputenc} 
\usepackage[T1]{fontenc}    
\usepackage{hyperref}       
\usepackage{url}            
\usepackage{booktabs}       
\usepackage{amsfonts}       
\usepackage{nicefrac}       
\usepackage{makecell}       
\usepackage{microtype}      
\usepackage{lipsum}		
\usepackage{graphicx}
\usepackage[numbers,sort&compress]{natbib}
\usepackage{doi}
\usepackage{caption}
\usepackage{subcaption} 
\usepackage{threeparttable} 
\usepackage{tabularray}
\usepackage{adjustbox}

\title{CoAtNet-DeepMoE: A Convolution--Attention Hybrid with DeepSeek Mixture-of-Experts for Parameter-Efficient Tomato Disease Classification}
  
\author{
Md Nadim Mahamood \\
Department of Computer Science and Engineering\\
Begum Rokeya University\\
Rangpur, Bangladesh \\
\texttt{nadim.cse.brur@gmail.com} \\

\and

Md Arif Shahriar \\
Department of Computer Science\\
Texas State University\\
San Marcos, Texas, USA \\
\texttt{arif.shahriar@txstate.edu} \\

\and

Md Shafi Ud Doula \\
Department of Information and Communication Technologies\\
Asian Institute of Technology\\
Pathum Thani, Thailand \\
\texttt{shafi.cse.brur@gmail.com} \\

\and

Kamrul Hasan \\
Department of Computer Science\\
Texas State University\\
San Marcos, Texas, USA \\
\texttt{kamrul.hasan@txstate.edu} \\
}

\date{}

\renewcommand{\shorttitle}{\textit{CoAtNet-DeepMoE} tomato disease classification}

\hypersetup{
pdftitle={A template for the arxiv style},
pdfsubject={q-bio.NC, q-bio.QM},
pdfauthor={David S.~Hippocampus, Elias D.~Striatum},
pdfkeywords={First keyword, Second keyword, More},
}

\begin{document}
\maketitle
\begin{abstract}
The world population is growing rapidly, and technology is improving in parallel. Meeting the huge demand for food for these 7 billion people not only depends on increasing food production but also on reducing food loss. Crop losses due to disease affect both the food supply and the financial and economic stability of a country. Tomatoes are among the top food-producing crops globally, and a significant portion of this production is lost due to disease. People have used Machine Learning techniques for feature extraction and early diagnosis of tomato diseases, and nowadays, Deep Learning-based models are widely used for disease recognition. However, most existing models are highly parameter-intensive, which increases the time required for training and inference. As a result, while lightweight models are more suitable for user-friendly applications, they often show a reduction in performance.
To balance performance and model size, we propose CoAtNet-DeepMoE, a Convolution–Attention hybrid architecture for rich feature extraction, further enhanced with a DeepSeek Mixture of Experts to substantially reduce the number of parameters without sacrificing accuracy. We evaluate our model on both balanced and imbalanced datasets from Kaggle and PlantVillage, demonstrating robustness and achieving 99.80\% accuracy, 99.80\% precision, 99.80\% recall, and 99.80\% F1-score on Kaggle, and 99.83\% accuracy, 99.85\% precision, 99.76\% recall, and 99.80\% F1-score on PlantVillage, representing state-of-the-art performance with only 2.47M parameters. The source code will be available at \url{https://github.com/nadimbrur/CoAt-MoE}.
\end{abstract}

\keywords{
DeepSeek MoE
\and 
Parameter optimization
\and 
Deep Learning 
\and 
Tomato disease
}

\section{Introduction}

The continuous growth of the global population has heightened the demand for food, and the Office of the Director of National Intelligence projects that the world population will reach 9.2 billion by 2040.\footnote{\url{https://www.dni.gov/files/ODNI/documents/assessments/GlobalTrends_2040.pdf}}
Furthermore, the Food and Agriculture Organization of the United Nations (FAO) estimates that global food production must increase by approximately 70\% by 2050 to meet future needs.\footnote{\url{https://www.fao.org/newsroom/detail/2050-A-third-more-mouths-to-feed/}} However, recent evidence shows that 2.3–2.9 billion people are unable to afford a healthy diet because their incomes are insufficient relative to the rising cost of nutritious foods \cite{stehl2025global}, highlighting a widening gap between food availability and economic access. To satisfy rising food demand, agricultural technologies are advancing rapidly; nevertheless, plant diseases and pests remain major obstacles. For example, the FAO reports that pests cause 20–40\% of global crop losses annually, imposing an estimated economic burden of about USD 290 billion.\footnote{\url{https://www.fao.org/newsroom/detail/New-standards-to-curb-the-global-spread-of-plant-pests-and-diseases/en}}

Tomatoes are one of the most regularly planted crops around for nutritional value and financial worth \cite{sharma2025deep}. Globally, it is the leading vegetable crop which production exceeds 180 million tons annually \cite{das2025deep}, contributing nearly one-sixth of total vegetable production \cite{abreu2008tomatoes, azeez2019bioactive, mustafa2025sustainable}. However, the farmer reported that tomato losses due to diseases were 64.71\% \cite{molelekoa2025quality}, and losses from virus diseases should be around 2 to 5\% annually \cite{sanchez2024estimating}. In addition, 80\% to 90\% of the disease on the plant appears on its leaves \cite{zhang2015plant}. Therefore, it is crucial to monitor plant leaves for signs of disease to control its spread through early recognition.

Tomato leaf diseases are caused by viruses, fungi, and bacteria, and they show several visual symptoms, including wilting, lesions, and leaf browning \cite{sharma2025deep}. Traditional laboratory-based diagnostic techniques such as Western blotting, enzyme-linked immunosorbent assay, and microarrays have been used to detect these pathogens \cite{yulita2023mobile}. However, in practical agricultural settings, disease detection still largely depends on manual visual inspection by experts, a time-consuming, subjective, and error-prone process \cite{sharma2025deep, xie2015detection, madufor2017detection}. In addition, monitoring large fields is tedious for farmers and often requires specific training, experience in recognizing disease symptoms \cite{blancard2012tomato}, and broad knowledge of multiple plant diseases. Furthermore, manual inspections lack scientific consistency because farmers’ skills and backgrounds differ, making the process less reliable \cite{trivedi2021early}.

Traditional machine learning (ML) and modern deep learning (DL) techniques have been extensively applied in tomato plant disease analysis. In the early years, traditional machine learning focused on recognizing plant diseases using image processing and feature-based data analysis \cite{choudhary2020feature}. For example, Geetha et al. \cite{geetha2020plant} developed a plant leaf disease detection system that preprocesses images, extracts key features using Histogram of Oriented Gradients and Gray Level Co-occurrence Matrix, and classifies diseases with the K-Nearest Neighbors (KNN) algorithm. Similarly, Basavaiah et al. \cite{basavaiah2020tomato} proposed a method for detecting tomato leaf diseases by fusing multiple features, Color Histograms, Hu Invariant Moments, Haralick texture features, and Local Binary Patterns to improve classification accuracy. Gadade et al. \cite{gadade2021machine} applied median filtering, extracted shape, color, and texture features, and evaluated multiple classifiers, Support Vector Machine, KNN, Naive Bayes, Decision Trees, and Linear Discriminant Analysis for both disease type detection and severity assessment to guide treatment. Most traditional machine learning-based models rely on manually extracted features, and their performance is heavily dependent on feature quality \cite{maurya2025rai, kamilaris2018deep}, leading to less accurate results compared to modern approaches \cite{javidan2024tomato}. Early models also required high-resolution images with controlled backgrounds; currently, research focuses on handling images captured under complex, real-world conditions \cite{joshi2025precision}.

Deep learning has achieved significant success in both computer vision and natural language processing applications, including agriculture (e.g., rice disease detection \cite{uddin2024e2etca}), due to its ability to model complex patterns in challenging environments. Similarly, deep learning is increasingly applied in tomato leaf disease analysis.

For instance, Wu et al. \cite{wu2024tomato} used a ResNet50 model with augmentation techniques to improve classification performance, although the model’s computational complexity was high. Hong et al. \cite{hong2020tomato} employed transfer learning to reduce training data requirements, time, and computational costs. Nine types of tomato leaves, including healthy leaves, were classified using five deep network architectures: ResNet50, Xception, MobileNet, ShuffleNet, and DenseNet121-Xception. DenseNet121-Xception achieved the highest accuracy of 97.10\% but had the largest number of parameters, while ShuffleNet achieved 83.68\% accuracy with far fewer parameters. Recently, concepts from natural language processing, such as transformer attention \cite{vaswani2017attention}, have been applied in computer vision tasks, including tomato leaf disease classification. For example, Chelladurai et al. \cite{chelladurai2025classification} segmented tomato leaf images using U-Net, extracted features with VGG-16, and classified diseases using a transductive Long Short-Term Memory (LSTM) network with attention, achieving very high accuracy under controlled conditions. However, these deep learning models typically involve millions of parameters, requiring substantial training time and computational resources, which makes deployment on low-end devices challenging. Reducing the number of parameters can also degrade performance, making it difficult to maintain high accuracy with lightweight models.

To maintain efficiency in both model size and performance, we propose CoAtNet-DeepMoE, a model that integrates convolution and attention mechanisms to capture rich features. It also employs the DeepSeek Mixture-of-Experts approach to reduce the number of parameters without compromising performance. The contributions of this study are summarized as follows:

\begin{itemize}
    \item \textbf{Efficient Architecture:} Designed a model that reduces parameters from 26.67 million to 2.47 million (an $\approx90.73\%$ reduction) by implementing the latest DeepSeek Mixture-of-Experts (MoE), replacing a traditional multi-layer perceptron (MLP), without sacrificing accuracy.
    \item \textbf{State-of-the-Art Performance:} Achieved 99.80\% accuracy, 99.80\% precision, 99.80\% recall, and 99.80\% F1-score on the Kaggle dataset (balanced) and 99.83\% accuracy, 99.85\% precision, 99.76\% recall, and 99.80\% F1-score on the PlantVillage dataset (imbalanced), demonstrating robustness across both balanced and imbalanced datasets.
    \item \textbf{Ablation Study:} Performed an ablation study comparing well-known convolutional and attention-based models, including CoAtNet-Base \cite{dai2021coatnet}, ResNet50 \cite{he2016deep}, and ViT-Tiny \cite{dosovitskiy2020image}, to assess inference time, parameter efficiency, and overall performance.
\end{itemize}

\section{Related Work}

\subsection{Machine learning-based}

Machine learning-based approaches for tomato leaf disease detection typically rely on handcrafted or color-based feature extraction methods, which are then fed into traditional classifiers. These models depend entirely on manually designed feature descriptors. For instance, Gadade et al. \cite{gadade2020tomato} proposed a segmentation-based system where infected regions were segmented and analyzed using color, texture, and shape features for classification and severity measurement. Among the various combinations of feature extraction methods and classifiers evaluated, the HOG + SVM model achieved the highest performance, yet the accuracy remained relatively low at 48.77\% across 45 combinations on 3000 PlantVillage images. Similarly, Joshi et al. \cite{joshi2024precision} introduced a GA-KNN framework, where morphological, statistical, and textural features of tomato leaves were extracted, and a Genetic Algorithm selected the optimal subset of features. This approach enabled KNN to achieve a significantly higher accuracy of 94.3\% using only 13 features on the PlantVillage dataset. In addition, Khan et al. \cite{khan2024automated} developed a system using Gray Level Co-occurrence Matrix (GLCM) and Scale-Invariant Feature Transform (SIFT) features with a quadratic SVM classifier, achieving 92.3\% accuracy with ten-fold cross-validation on a 2700-image, nine-class dataset, demonstrating reliable multiclass classification.

Recent studies increasingly combine deep learning feature extraction with traditional machine learning classifiers. Imam et al. \cite{imam2024transfer} proposed a hybrid MobileNet–SVM approach, where features extracted from MobileNet were classified by an SVM, achieving 99.37\% overall accuracy across nine disease classes, outperforming previous methods. Similarly, Terziouglu et al. \cite{terziouglu2025comparative} evaluated 21 deep learning models for tomato disease detection on a 6414-image dataset. The top-performing combination used EfficientNet-b0 features with Chi-Square feature selection and a Fine KNN classifier, achieving 92.0\% accuracy, highlighting the effectiveness of deep feature extraction combined with traditional classifiers for robust disease identification.

\subsection{Deep learning-based}
Deep learning approaches are increasingly used in tomato leaf disease detection for automatically learning hierarchical image features. Major directions include CNNs, attention-based models, and hybrid approaches, which outperform traditional machine learning in accuracy and robustness.

CNN \cite{o2015introduction} focuses on extracting spatial features from images using various filters to identify shapes, textures, and other patterns for classification. It is prominently used in tomato leaf disease detection. For example, Assaduzzaman et al. \cite{assaduzzaman2025xse} developed XSE-TomatoNet, an enhanced EfficientNetB0 with Squeeze-and-Excitation (SE) blocks and multi-scale feature fusion, achieving 98.83\% test accuracy, outperforming MobileNet and VGG19. Since it is based on EfficientNetB0 (~6.0 million parameters), the total parameters are at least 6 million, with additional parameters from SE blocks and multi-scale fusion. Another study \cite{abdullah2024deep} employed YOLOv8s for tomato leaf disease detection on the PlantVillage dataset, attaining the highest mAP of 92.5\% and a fast inference speed of 121.5 FPS, outperforming YOLOv5 and Faster R-CNN in both accuracy and real-time detection. The estimated number of parameters for YOLOv8s is approximately 11.0 million. In addition, Prabhasha et al. \cite{prabhasha2025tomato} used a transfer learning-based approach with pre-trained models (CNN, AlexNet, ResNet, InceptionV3, VGG-16) for tomato leaf disease detection, with VGG-16 achieving the highest accuracy of 93.7\%, demonstrating efficient classification and reduced training time through transfer learning.

Hybrid models that combine CNN and attention mechanisms are increasingly popular for tomato leaf disease detection. For instance, Zhao et al. \cite{zhao2021tomato} proposed a deep CNN with residual blocks and attention modules, achieving 96.81\% accuracy. Similarly, Karthik et al. \cite{karthik2020attention} employed two architectures, residual CNN and residual CNN with attention, on the PlantVillage dataset, achieving 98\% validation accuracy using 5-fold cross-validation. Additionally, Sunil et al. \cite{sunil2023tomato} proposed a Multilevel Feature Fusion Network (MFFN) with ResNet50 and an Adaptive Attention Mechanism (channel, spatial, and pixel attention), achieving 99.88\% training and validation accuracy and 99.83\% external test accuracy, along with a pesticide recommendation module based on detected diseases.

Machine learning–based classification suffers from a lack of generalizability and depends heavily on handcrafted features \cite{uddin2024e2etca}. In contrast, deep learning achieves highly prominent performance in CNNs and attention-based models, but these come with a huge number of parameters, leading to high costs, long training times, large GPU power requirements, increased energy consumption, and a higher carbon footprint. Reducing parameters, however, can hinder performance. To address this, we propose CoAtNet-DeepMoE, which utilizes convolution and attention to capture fine-grained features and incorporates an advanced version of Mixture-of-Experts, DeepSeek-MoE, developed by DeepSeek, that reduces the number of parameters by splitting the large network into smaller subparts with a shared space for all. This implementation reduces the number of parameters from 26.67 million to 2.47 million, achieving approximately 90.73\% reduction, without any loss in classification accuracy.

\begin{figure*}[!htbp]
\centering
\includegraphics[width=\linewidth]{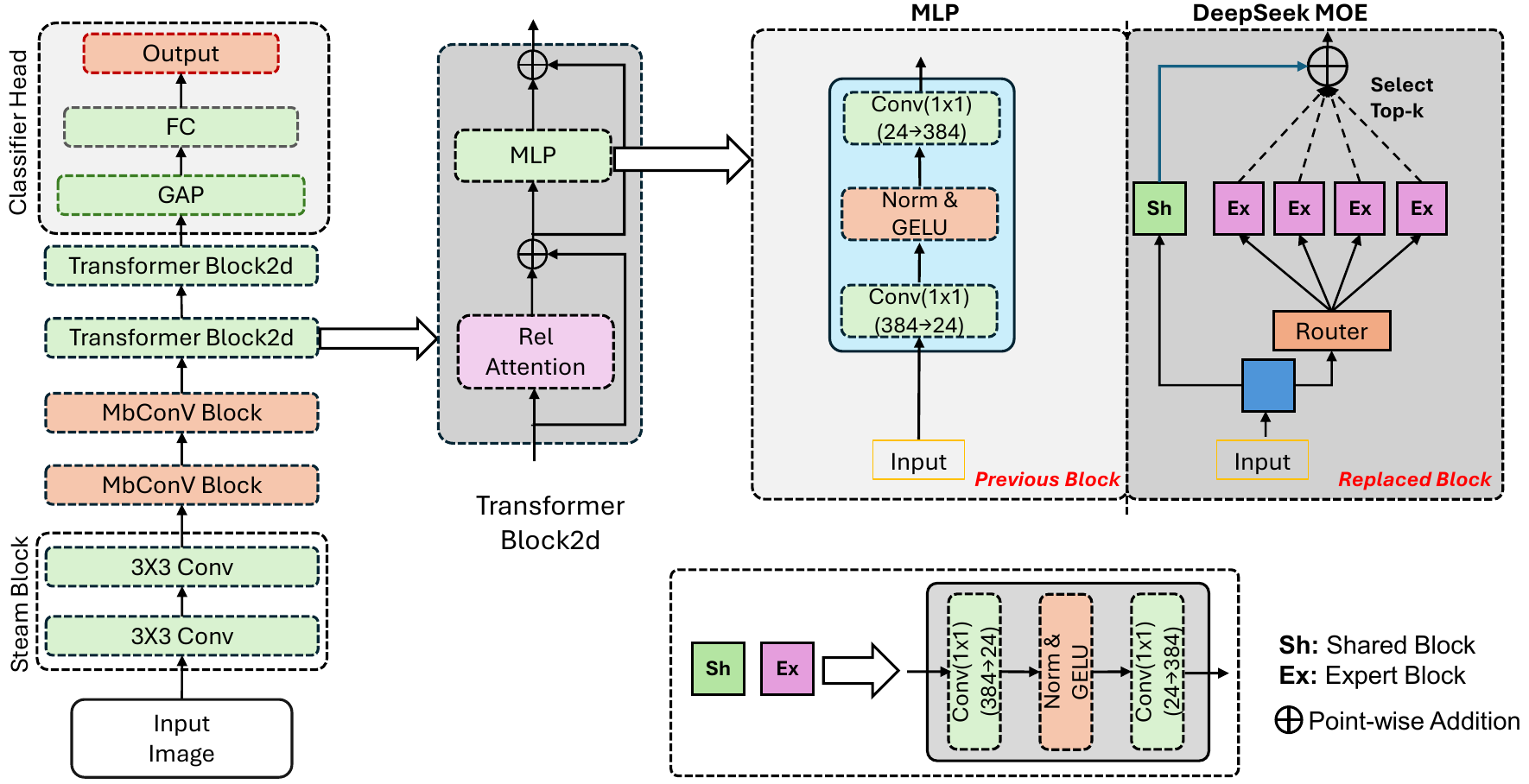}
\caption{Overview of the proposed CoAtNet-DeepMoE architecture. The model integrates convolutional layers for local feature extraction, MbConv and SE blocks for channel-wise enhancement, and attention layers for global feature modeling. A parameter-efficient DeepSeek Mixture-of-Experts replaces the traditional MLP, where the router selects the top-k experts ($2$ in this case) along with a shared expert for feature processing. The selected experts specialize in capturing disease-specific patterns, enabling the model to focus on the most relevant features for each input. The shared expert, on the other hand, preserves common information across all classes (e.g., common characteristics), ensuring stable and generalizable representations. This selective expert activation enhances feature representation while greatly reducing computational cost and overall parameter count ($26.67\,M$ to $2.47\,M$, $\approx 90.73\%$).}
\label{fig:1_architecture}
\end{figure*}

\subsection{Methods}
The proposed CoAtNet-DeepMoE model combines the strengths of convolutional networks, attention mechanisms present in CoAtNet \cite{dai2021coatnet}, and Mixture-of-Experts (MoE) \cite{mu2025comprehensive} to achieve an efficient and high-performance architecture for tomato leaf disease classification. The methodology is composed of several key stages that progressively transform the input image into a compact, discriminative representation suitable for final classification.

\subsubsection{Stem Block}
The stem block is the first stage of the proposed model, responsible for extracting low-level spatial features from the input image. It consists of two consecutive $3 \times 3$ convolution layers with strides $2$ and $1$, respectively, along with a Layer Normalization applied between them. This block captures essential early information such as edges, textures, and color transitions, while also reducing the spatial resolution to prepare the image for deeper feature extraction.

Given an input image $X \in \mathbb{R}^{224 \times 224 \times 3}$, the stem block operation can be formulated as:
\begin{equation}
    X_s = \text{Conv}_{3 \times 3}\big(\text{Norm}(\text{Conv}_{3 \times 3}(X))\big),
\label{eq:stem_block}
\end{equation}
where the first $\text{Conv}_{3 \times 3}(\cdot)$ projects the feature map to $\mathbb{R}^{112 \times 112 \times 32}$, and the second $\text{Conv}_{3 \times 3}(\cdot)$ further projects it to $\mathbb{R}^{112 \times 112 \times 64}$. This stage efficiently reduces spatial dimensions while preserving important visual information needed for subsequent MbConv and Transformer layers.

\subsubsection{MbConv Block}

The MbConv block \cite{sandler2018mobilenetv2} consists of depthwise convolution \cite{laurent2014rigid} to capture spatial interactions. In equation form, it can be written as:

\begin{equation}
X_{\text{MbConv}} =
\begin{aligned}[t]
&\underbrace{\text{Conv}_{1\times1} (\text{AvgPool}_{2\times2}(X_s))}_{\text{Skip connection}} + 
&\underbrace{\text{Conv}_{1\times1} \Big(
\text{SE} \big(
\text{DepConv}_{3\times3} \big(
\text{Conv}_{1\times1} (\text{AvgPool}_{2\times2}(X_s))
\big)
\big)
\Big)}_{\text{Main path}}
\end{aligned}
\end{equation}

This block has two paths: the skip connection performs spatial downsampling by a factor of $2$ followed by a $1 \times 1$ convolution to project features into the desired dimension. The main path consists of downsampling, a $1 \times 1$ convolution that expands the channel dimension by 4×, a depthwise convolution ($DepthConv_{3 \times 3}(\cdot)$) to capture spatial interactions, and a Squeeze-and-Excitation (SE) module, which adaptively recalibrates channel-wise feature responses by emphasizing important channels and suppressing less useful ones without affecting the feature dimensions. Finally, another $1 \times 1$ convolution projects the features to match the skip connection dimensions.

After the first MbConv block, the output feature has dimensions $\mathbb{R}^{56 \times 56 \times 96}$, and after the second block, the output feature has dimensions $\mathbb{R}^{28 \times 28 \times 192}$.

\subsubsection{Transformer2d Block}
\textbf{Relative Attention:}
CoAtNet \cite{dai2021coatnet} employs a variant of the transformer attention mechanism known as relative attention \cite{raffel2020exploring}, which is input-independent and eliminates both parameter sharing across layers and the need for a bucketing mechanism.
The first portion of Transformer2d consists of two parallel paths: the skip connection, where the input is downsampled by a factor of 2 and passed through a $1 \times 1$ convolution to expand the number of channels; and the main path, where the downsampled input is processed by the relative attention module. This module captures global receptive fields and models long-range spatial dependencies across the leaf surface, such as vein patterns, texture irregularities, and spot distributions, which are essential for accurate disease identification. It is outlined as:

\begin{align}
X_{\text{attention}} &=
\underbrace{\text{Rel\_attention}\big(\text{AvgPool}_{2\times2}(X_{\text{MbConv}})\big)}_{\text{Main path}} \notag +
\underbrace{\text{Conv}_{1\times1}\big(\text{AvgPool}_{2\times2}(X_{\text{MbConv}})\big)}_{\text{Skip connection}}
\end{align}

\textbf{DeepSeek MoE:} We replace the traditional multi-layer perceptron (MLP) with DeepSeek MoE \cite{dai2024deepseekmoe} for parameter optimization. DeepSeek MoE is an improved version of the Mixture-of-Experts (MoE) \cite{mu2025comprehensive} framework, in which the network is divided into several smaller subnetworks called experts. Unlike standard MLPs, which have a large number of parameters, MoE reduces the parameter count by activating only a subset of experts for each input, determined dynamically by a gating router.

DeepSeek MoE further introduces a shared expert, which provides information common to all inputs. While the top-$k$ experts are dynamically selected for each input via the gating function, the shared expert is always activated, ensuring essential information flows across all input types. For the input $X_{\text{attention}}$, the operation can be formulated as:

\begin{equation}
X_{\text{DeepSeek MoE}} = S(x) + \sum_{i \in \text{TopK}(G(x), 2)} g_i \cdot E_i(x)
\end{equation}

where $E_i(\cdot)$ denotes the output of the $i$-th expert, $S(\cdot)$ is the output of the shared expert, $G(\cdot)$ represents the gating function producing scores $g_i$, and $\text{TopK}(G(x), 2)$ indicates the indices of the top-2 selected experts. Each expert first reduces the input channels by a factor of four and then projects them back to the original dimension using $1 \times 1$ convolutions.

The output of DeepSeek MoE is then combined with the original input via a residual connection:

\begin{equation}
X_{\text{MLP}} = \underbrace{X_{\text{DeepSeek MoE}}}_{\text{Main path}} + \underbrace{X_{\text{attention}}}_{\text{Skip connection}}
\end{equation}

Finally, the model includes two Transformer2D stages. The first outputs a feature map of size $\mathbb{R}^{14 \times 14 \times 384}$, and the second outputs a feature map of size $\mathbb{R}^{7 \times 7 \times 768}$, which we denote as $X_{\text{MLP}}$ before passing it to the classifier head.

\subsubsection{Classifier head}

The classifier head consists of a Global Average Pooling (GAP) layer that converts the 2D feature map into a 1D feature vector. This vector is then passed through a fully connected (FC) layer to project it into the desired number of output classes. Finally, a softmax activation is applied to compute the class probabilities. Mathematically, this can be expressed as:

\begin{equation}
X_{\text{class}} = \text{FC} \big( \text{GAP}(X_{\text{MLP}}) \big)
\end{equation}

where $\text{GAP}(\cdot)$ denotes the Global Average Pooling operation, which converts the feature map $X_{\text{MLP}} \in \mathbb{R}^{7 \times 7 \times 768}$ into a 1D feature vector of size $768$, and $\text{FC}(\cdot)$ projects this vector into the number of output classes, here $10$. The use of DeepSeek MoE reduces the total number of parameters from 26,674,252 to 2,473,316 ($\approx$90.73\% reduction) without any loss in accuracy. The overall architecture is illustrated in Fig.~\ref{fig:1_architecture}, and as shown in Table~\ref{tab:parameters_comparison}, the block-wise parameter comparison highlights that CoAt-DeepMoE substantially reduces the number of parameters in both transformer stages, while the stem and classifier head remain unchanged.

\begin{table}[h!]
\centering
\small
\renewcommand{\arraystretch}{1.4}
\caption{Block-wise parameter comparison between CoAt-Base and CoAt-DeepMoE.}
\label{tab:parameters_comparison}
\begin{tabular}{lrr}
\hline
\textbf{Module} & \textbf{CoAt-Base} & \textbf{CoAt-DeepMoE} \\
\hline
Stem            &         19,360 &         19,360 \\ \hline
MbConv-0        &        236,768 &         83,648 \\ \hline
MbConv-1        &      1,410,720 &        208,416 \\ \hline
Transformer2D-2 &     12,145,374 &        437,174 \\ \hline
Transformer2D-3 &     12,852,804 &      1,715,492 \\ \hline
Classifier Head &          9,226 &          9,226 \\ \hline
\textbf{Total}  & \textbf{26,674,252} & \textbf{2,473,316} \\ \hline
\end{tabular}
\end{table}

\subsubsection{Dataset}

\begin{table*}[!t]
\centering
\caption{Corrected class distribution comparison between Kaggle and PlantVillage tomato datasets (instances and ratios).}
\resizebox{\linewidth}{!}{%
\begin{tabular}{|l|c|c|c|c|c|c|c|c|c|c|c|}
\hline
\textbf{Dataset} & BS  & EB & LB & LM & SLS & SE & TS & YLCV & MV & Healthy & \textbf{Total} \\
\hline
Kaggle& 1100 (0.10) & 1100 (0.10) & 1100 (0.10) & 1100 (0.10) & 1100 (0.10)& 1100 (0.10) & 1100 (0.10) & 1100 (0.10) & 1100 (0.10) & 1100 (0.10)
& 11000 (1.00) \\ \hline

PlantVillage& 2127 (0.12) & 1000 (0.06) & 1909 (0.11) & 952 (0.05) & 1771 (0.10)& 1676 (0.09) & 1404 (0.08) & 5357 (0.30) & 373 (0.02) & 1591 (0.09)& 18160 (1.00) \\ \hline

\textbf{Total}& 3227 (0.11) & 2100 (0.07) & 3009 (0.10) & 2052 (0.07) & 2871 (0.09) & 2776 (0.10) & 2504 (0.09) & 6457 (0.22) & 1473 (0.05) & 2691 (0.09) & 29160 (1.00) \\ \hline

\multicolumn{12}{l}{\makecell[l]{Bacterial spot: BS; Leaf Mold: LM; Early blight: EB; Late blight: LB; Septoria leaf spot: SLS; Spider mites: SE; Target Spot: TS; Yellow Leaf Curl Virus: YLCV; Mosaic virus: MV}} \\
\end{tabular}
}
\label{tab:dataset}
\end{table*}

\begin{figure*}[h!]
\centering
\begin{minipage}{0.19\linewidth}
    \centering
    \includegraphics[width=\linewidth]{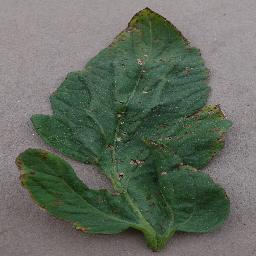}
    \subcaption{Bacterial spot}
    \label{fig:dataset2a}
\end{minipage}
\hfill
\begin{minipage}{0.19\linewidth}
    \centering
    \includegraphics[width=\linewidth]{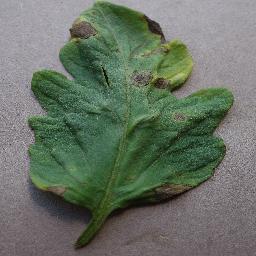}
    \subcaption{Early blight}
    \label{fig:dataset2b}
\end{minipage}
\hfill
\begin{minipage}{0.19\linewidth}
    \centering
    \includegraphics[width=\linewidth]{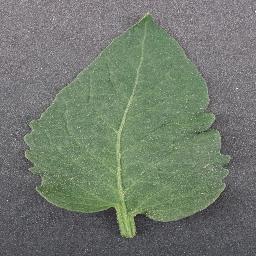}
    \subcaption{Healthy}
    \label{fig:dataset2c}
\end{minipage}
\hfill
\begin{minipage}{0.19\linewidth}
    \centering
    \includegraphics[width=\linewidth]{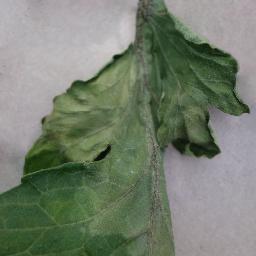}
    \subcaption{Late blight}
    \label{fig:dataset2d}
\end{minipage}
\hfill
\begin{minipage}{0.19\linewidth}
    \centering
    \includegraphics[width=\linewidth]{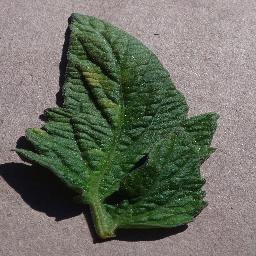}
    \subcaption{Leaf Mold}
    \label{fig:dataset2e}
\end{minipage}
\\
\begin{minipage}{0.19\linewidth}
    \centering
    \includegraphics[width=\linewidth]{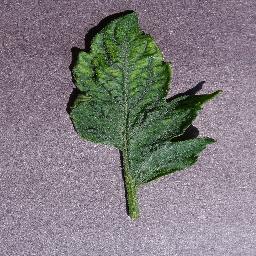}
    \subcaption{Mosaic virus}
    \label{fig:dataset2f}
\end{minipage}
\hfill
\begin{minipage}{0.19\linewidth}
    \centering
    \includegraphics[width=\linewidth]{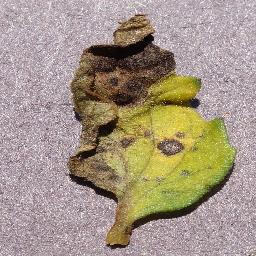}
    \subcaption{Septoria leaf spot}
    \label{fig:dataset2g}
\end{minipage}
\hfill
\begin{minipage}{0.19\linewidth}
    \centering
    \includegraphics[width=\linewidth]{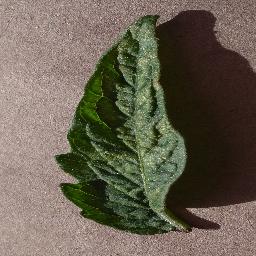}
    \subcaption{Two-spotted spider}
    \label{fig:dataset2h}
\end{minipage}
\hfill
\begin{minipage}{0.19\linewidth}
    \centering
    \includegraphics[width=\linewidth]{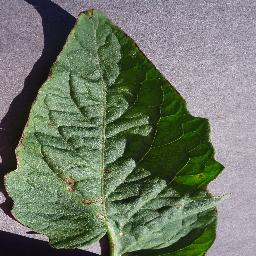}
    \subcaption{Target spot}
    \label{fig:dataset2i}
\end{minipage}
\hfill
\begin{minipage}{0.19\linewidth}
    \centering
    \includegraphics[width=\linewidth]{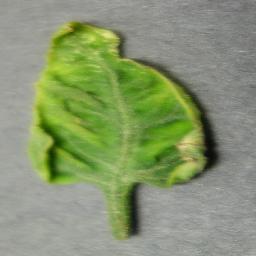}
    \subcaption{Yellow leaf curl virus}
    \label{fig:dataset2j}
\end{minipage}
\caption{Examples of representative images from each class in the tomato leaf disease datasets, illustrating the visual diversity and key characteristics.}
\label{fig:dataset_example}
\end{figure*}

The tomato disease dataset used in this study is constructed by combining images from the Kaggle repository\footnote{https://www.kaggle.com/datasets/kaustubhb999/tomatoleaf} and the PlantVillage benchmark dataset\footnote{https://www.kaggle.com/datasets/abdallahalidev/plantvillage-dataset}, resulting in a total of 29,160 images across ten classes. While the Kaggle dataset maintains a perfectly uniform distribution with 1,100 samples per class, the PlantVillage dataset is significantly imbalanced. In particular, Yellow Leaf Curl Virus (YLCV) is the dominant class, contributing 5,357 images (0.30), whereas classes such as Mosaic Virus (MV), with 373 images (0.02), and Leaf Mold (LM), with 952 images (0.05), are notably underrepresented. When combined, the overall dataset still exhibits class imbalance, with YLCV remaining the largest class (6,457 images; 0.22) and MV becoming the least represented class (1,473 images; 0.05). Such an imbalance can negatively affect model learning by biasing predictions toward frequent classes while reducing sensitivity to rare diseases. To mitigate this issue, we apply targeted data augmentation, such as rotations, flips, and color jittering, to improve class uniformity and enhance the model's ability to generalize across all disease categories. The distribution of the Kaggle dataset, the PlantVillage dataset, and the combined dataset is shown in Table \ref{tab:dataset}. Also, an image example for each class of the dataset is shown in Figure \ref{fig:dataset_example}.

\subsubsection{Training details}
The proposed CoAtNet-DeepMoE model was trained for $300$ epochs with a batch size of $32$, using input images resized to $224 \times 224 \times 3$. To stabilize training and improve robustness, a Cosine Annealing learning rate scheduler \cite{pytorch_cosineannealinglr} was employed. The model was trained using Cross-Entropy loss with label smoothing \cite{szegedy2016rethinking} set to 0.1 to reduce overconfidence in predictions. Label smoothing works by assigning a slightly lower probability to the correct class and distributing the remaining probability across the other classes. For instance, with a smoothing value of 0.1, the correct class receives a probability of 0.9, while the remaining 0.1 is distributed among the other classes. This technique helps improve generalization and reduces overfitting. Additional training details, including optimizer, weight decay, early stopping, and hardware specifications, are summarized in Table \ref{tab:training_details}.

\subsubsection{Performance evaluation metrics}

\begin{table}[!t]
\centering
\caption{Training configuration details used for model training.}
\renewcommand{\arraystretch}{1.3}
\begin{tabular}{l l}
\hline
\textbf{Parameter} & \textbf{Value} \\ \hline
Learning Rate (LR) & $1 \times 10^{-5}$ \\ \hline
Loss Function & CE (label\_smoothing = 0.1) \\ \hline
Optimizer & AdamW (weight\_decay = $1 \times 10^{-4}$) \\ \hline
LR Scheduler & CosineAnnealingLR ($T_{\text{max}} = 300$) \\ \hline
Stopping Patience & 50 epochs \\ \hline
Batch Size & 32 \\ \hline
Input Image Size & (224, 224, 3) \\ \hline
Activation Function & Softmax \\ \hline
Hardware & NVIDIA GeForce RTX 3090 GPU\\ \hline
\end{tabular}
\label{tab:training_details}
\end{table}

We consider four evaluation metrics, along with the number of parameters of the model, to validate the performance of the proposed approach. These metrics include Accuracy, Precision, Recall, and F1-score. Accuracy represents the ratio of correctly classified instances to the total number of predictions. Precision measures the proportion of correctly identified positive samples among all predicted positive samples. For example, when computing the precision of the healthy class, we treat healthy as the positive class and the remaining classes as negative; thus, precision indicates how many predicted healthy instances are actually healthy. Recall, in contrast, measures how many true positive samples are correctly identified among all actual positive samples. Following the same example, the recall of the healthy class reflects how many truly healthy instances are correctly classified as healthy. The F1-score is then computed as the harmonic mean of precision and recall, providing a balanced assessment of model performance.

Since one of our datasets is imbalanced, we report the macro-averaged Precision, Recall, and F1-score to ensure equal contribution of each class to the final evaluation. Table~\ref{tab:evaluation_metric} presents the mathematical formulations of these metrics.

\begin{table} [!t]
\centering
\caption{Macro-averaged metrics used to evaluate the proposed CoAtNet-DeepMoE}
\label{tab:evaluation_metric}
\begin{tabular}{lc} \hline
Metric & Expression  \\
\hline \\
Accuracy &
$\displaystyle \frac{\sum_{i=1}^{K} \mathrm{TP}_i}{\sum_{i=1}^{K} (\mathrm{TP}_i + \mathrm{FP}_i + \mathrm{FN}_i)}$\\ \\
Recall &$\displaystyle \frac{1}{K}\sum_{i=1}^{K} \frac{\mathrm{TP}_i}{\mathrm{TP}_i+\mathrm{FN}_i}$\\ \\
Precision &$\displaystyle \frac{1}{K}\sum_{i=1}^{K} \frac{\mathrm{TP}_i}{\mathrm{TP}_i+\mathrm{FP}_i}$\\ \\
F1-score  &$\displaystyle\frac{1}{K}\sum_{i=1}^{K} \frac{2\,\mathrm{TP}_i}{2\,\mathrm{TP}_i+\mathrm{FP}_i+\mathrm{FN}_i}$\\ \\\hline
\end{tabular}
\end{table}

\section{Experiment \& Results}
\subsection{Comparison with SOTA}
\subsubsection{Performance on the Kaggle dataset and compare with the SOTA models:}

\begin{table*}[htbp]
\centering
\caption{Performance comparison of state-of-the-art (SOTA) models on the Kaggle tomato disease dataset. ``M'' denotes the number of trainable parameters in millions. Accuracy, Precision, Recall, and F1-score are reported in percentage (\%) form. Bold values indicate the best performance among the compared models.}
\label{tab:performance_comparison_Kaggle}
\renewcommand{\arraystretch}{1.4}
\resizebox{\textwidth}{!}{
\begin{tabular}{c c c l c c c c}
\hline
\textbf{SL} & \textbf{Reference} & \textbf{Parameters} & \textbf{Model Name} &\textbf{Accuracy} & \textbf{Precision} & \textbf{Recall} & \textbf{F1-score} \\\hline

1 & \cite{abdullah2024deep} & 11.00 M & YOLOv8s& 93.2 & 90.3 & -- & -- \\ \hline
2 & \cite{yulita2023mobile} & 32.29 M & DenseNet& 95.4 & 95.7 & 95.4 & 95.4 \\ \hline
3 & \cite{wu2024tomato} & 23.50 M & ResNet50& 92.4 & -- & -- & -- \\ \hline
4 & \cite{wu2024tomato} & 23.56 M & SDE-ResNet50& 98.0 & 98.0 & -- & 98.0 \\ \hline
5 & \cite{wu2024tomato} & 54.32 M & Inception-ResNet& 98.2 & 98.23 & -- & 98.2 \\ \hline
6 & \cite{cengil2022hybrid} & 200.00 M &\begin{tabular}[c]{@{}l@{}}AlexNet + ResNet50\\+ VGG16 + QSSVM\end{tabular}& 98.3 & -- & -- & -- \\ \hline
7 & \cite{am2024improved} & 143.30 M &\begin{tabular}[c]{@{}l@{}}VGG16 + NASNet\end{tabular}& 98.7 & 97.9 & 98.6 & 98.6 \\\hline
8 & \textbf{Our} & \textbf{2.47 M} & \textbf{CoAtNet-DeepMoE}& \textbf{99.80} & \textbf{99.80} & \textbf{99.80} & \textbf{99.80} \\ \hline
\end{tabular}
}
\end{table*}

\begin{figure*}[!t]
\centering
\begin{minipage}{0.90\linewidth}
    \centering
    \includegraphics[width=\linewidth]{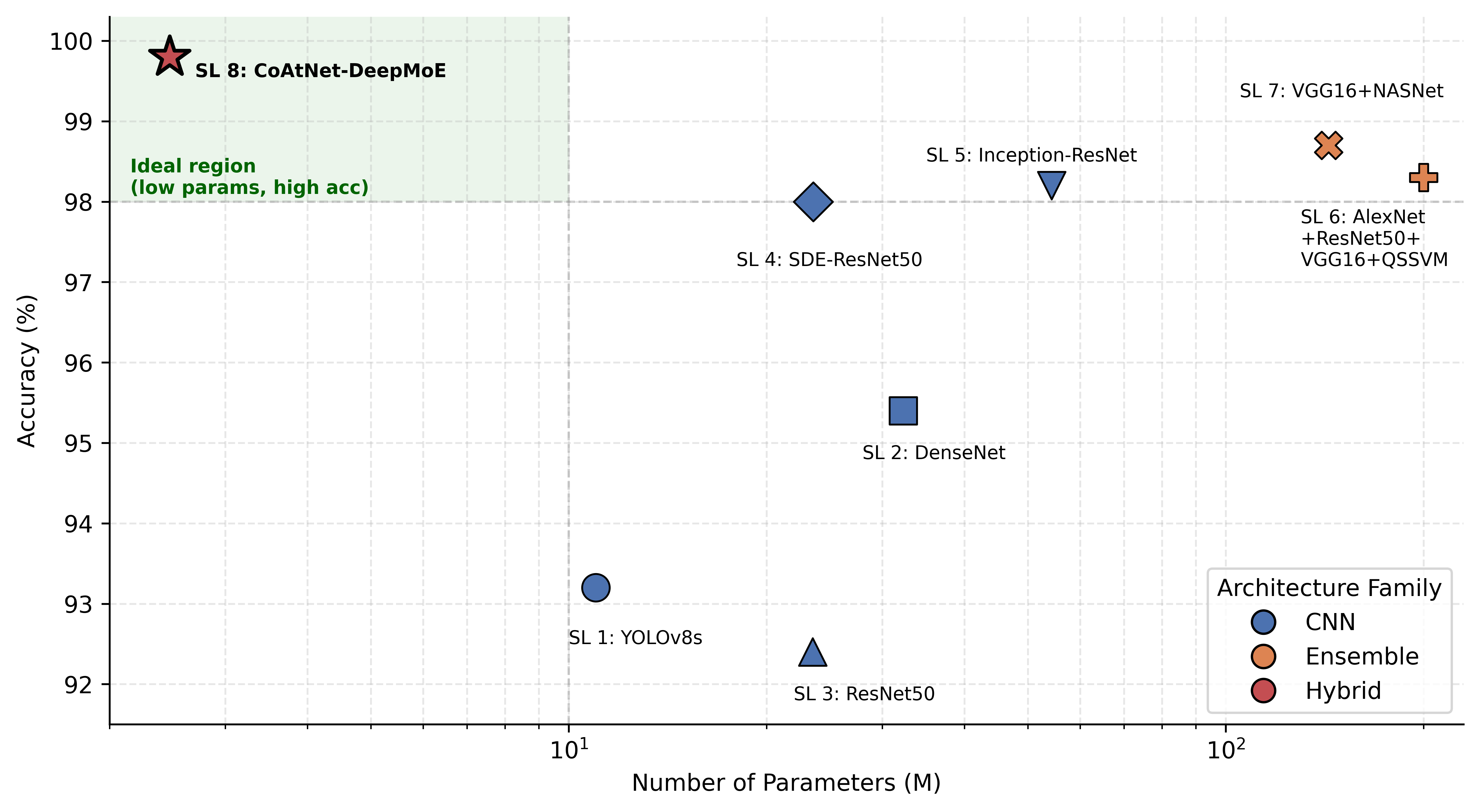}
    \subcaption{Kaggle}
    \label{fig:3a_Kaggle}
\end{minipage}
\hfill
\begin{minipage}{0.90\linewidth}
    \centering
    \includegraphics[width=\linewidth]{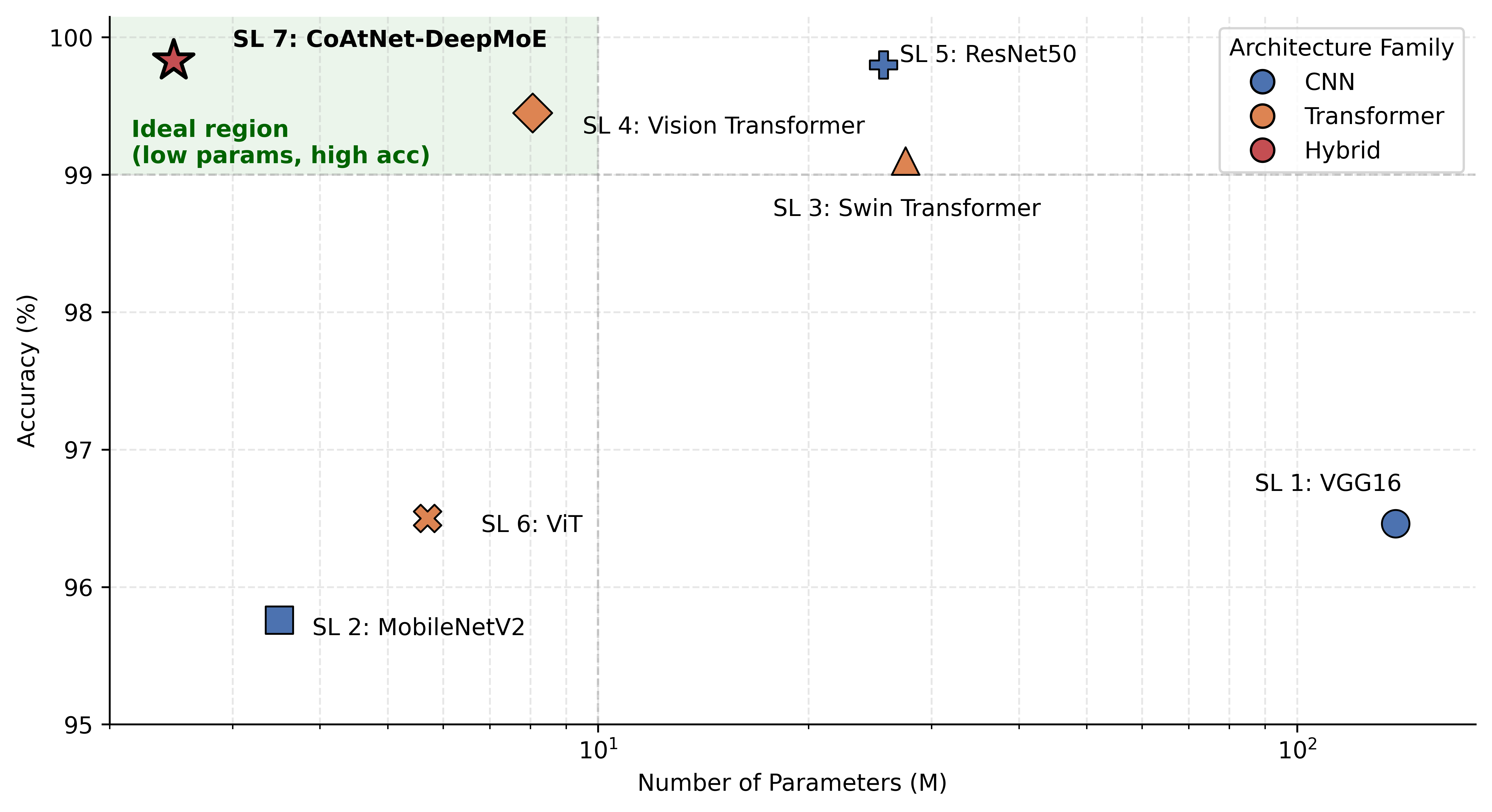}
    \subcaption{PlantVillage}
    \label{fig:3b_Plant_village}
\end{minipage}
\caption{2D comparison of SOTA models on (a) the Kaggle dataset and (b) the PlantVillage dataset. Each plot presents model accuracy against parameter count (in millions), illustrating how different architectures balance predictive performance and model complexity.}
\label{fig:SOTA_comparison}
\end{figure*}

\begin{figure*}[!t]
\centering
\begin{minipage}{0.495\linewidth}
    \centering
    \includegraphics[width=\linewidth]{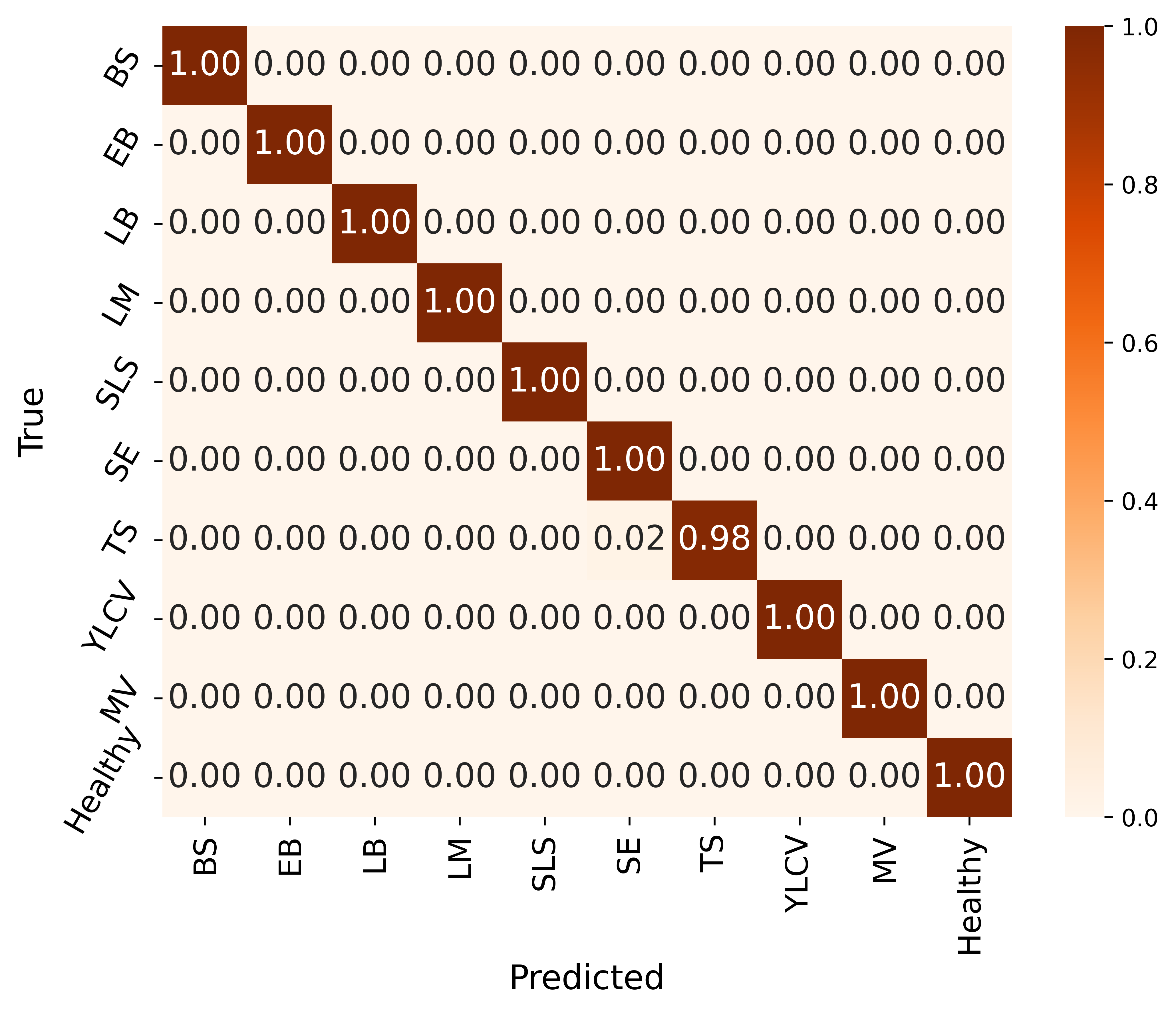}
    \subcaption{Kaggle}
    \label{fig:4a_Kaggle_confusion_matrix}
\end{minipage}
\hfill
\begin{minipage}{0.495\linewidth}
    \centering
    \includegraphics[width=\linewidth]{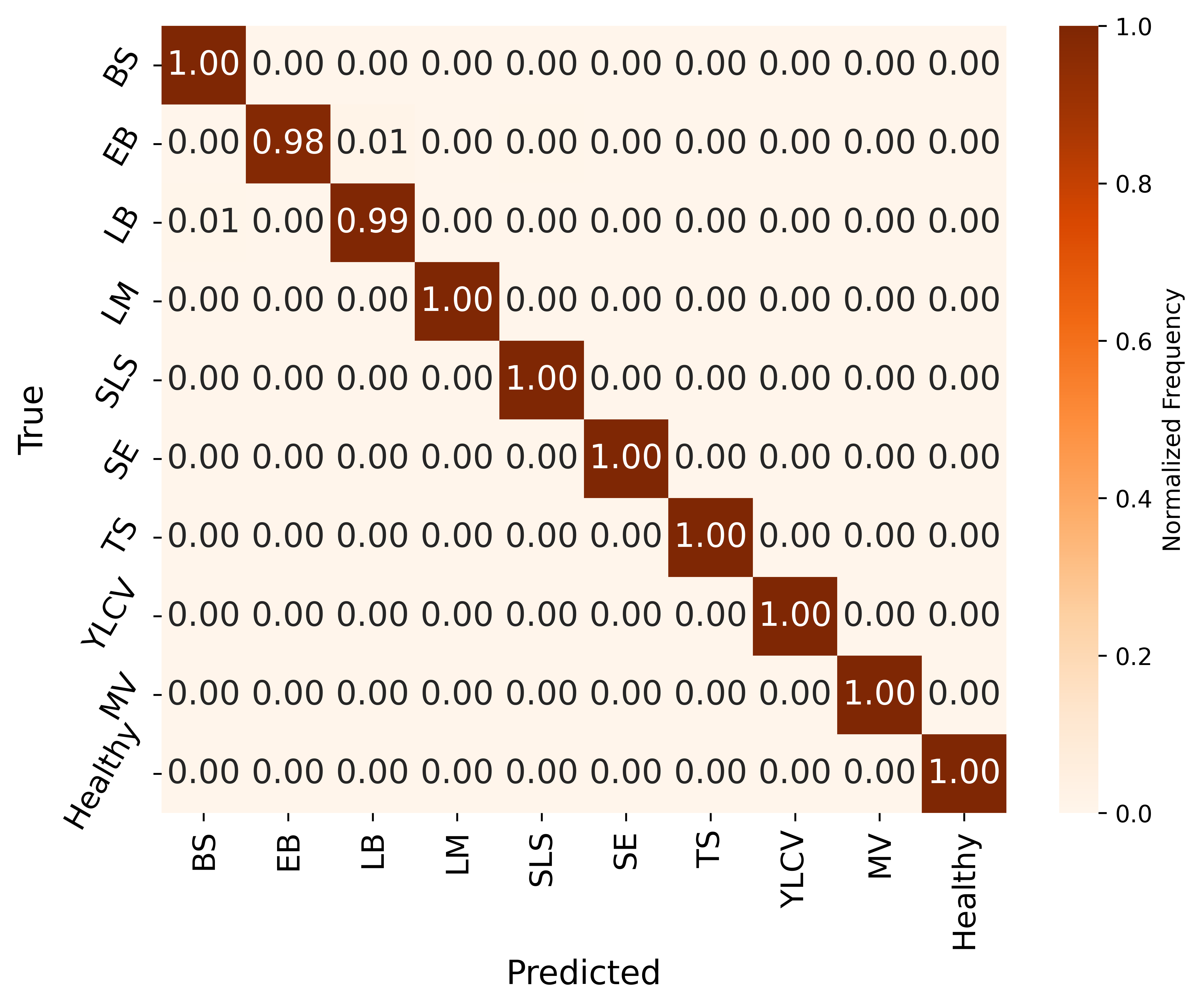}
    \subcaption{PlantVillage}
    \label{fig:4b_Plant_village_confusion_matrix}
\end{minipage}
\caption{Normalized confusion matrices of the proposed CoAtNet-DeepMoE model on (a) the Kaggle tomato leaf disease dataset and (b) the PlantVillage tomato dataset.}
\label{fig:confusion_matrices_kaggle_pv}
\end{figure*}

Our proposed CoAtNet-DeepMoE model was evaluated on the Kaggle dataset and compared with several SOTA models, as presented in Table~\ref{tab:performance_comparison_Kaggle}. The comparison includes the number of parameters, Accuracy, Precision, Recall, and F1-score. Model parameter counts were taken directly from the corresponding papers; when unavailable, we used the smallest official model variant and computed the parameters ourselves. The dataset was split according to the partition provided by the original authors, and the confusion matrix of CoAtNet-DeepMoE for all classes is shown in Figure~\ref{fig:4a_Kaggle_confusion_matrix}. As observed, CoAtNet-DeepMoE achieves the best performance across all metrics. The closest model in terms of Accuracy is VGG16+NASNet \cite{am2024improved}, which reaches 98.7\% accuracy; however, it requires approximately 58 times more parameters than CoAtNet-DeepMoE (143.30M vs. 2.47M).

On the other hand, \cite{abdullah2024deep} is the closest in terms of parameter count, with 11.00M parameters, approximately 4.5 times more than our model, yet its performance is 6.6\% lower in Accuracy and 9.5\% lower in Precision. These results demonstrate the strong feature-extraction capability of CoAtNet-DeepMoE while significantly reducing the number of parameters through the integration of DeepSeek-MoE. Figure~\ref{fig:3a_Kaggle} further illustrates how CoAtNet-DeepMoE surpasses existing models in both Accuracy and model size.

\subsubsection{Performance on the PlantVillage dataset and compare with the SOTA models:}

Our proposed CoAtNet-DeepMoE model was further evaluated on the PlantVillage dataset and compared with several SOTA models, as shown in Table~\ref{tab:performance_comparison_Plant_village}. Similar to the earlier evaluation, the comparison includes the number of parameters, Accuracy, Precision, Recall, and F1-score. The parameter counts of the baseline models were followed as reported in the Kaggle implementations. For fairness, we used the same train–test split provided by \cite{karimanzira2025context}, as their split configuration is widely adopted and yields the most consistent results. The confusion matrix of CoAtNet-DeepMoE for all classes is presented in Figure~\ref{fig:4b_Plant_village_confusion_matrix}.

\begin{table*}[h!]
\centering
\caption{Performance comparison of state-of-the-art (SOTA) models on the PlantVillage dataset. ``M'' denotes the number of parameters in millions. Accuracy, Precision, Recall, and F1-score are reported in percentage (\%) form. Bold values indicate the best performance among the compared models.}
\label{tab:performance_comparison_Plant_village}
\renewcommand{\arraystretch}{1.4}
\resizebox{\textwidth}{!}{
\begin{tabular}{c c c c c c c c}\hline
\textbf{SL} & \textbf{Reference} & \textbf{Parameters} & \textbf{Model Name} & \textbf{Accuracy} & \textbf{Precision} & \textbf{Recall} & \textbf{F1-score} \\\hline
1 & \cite{chen2024using} & 138.36 M & VGG16 & 96.46 & -- & -- & -- \\\hline
2 & \cite{chen2024using} & 3.50 M & MobileNetV2 & 95.76 & -- & -- & -- \\\hline
3 & \cite{chen2024using} & 27.53 M & Swin Transformer & 99.10 & -- & -- & -- \\\hline
4 & \cite{chen2024using} & 8.06 M & Vision Transformer & 99.45 & -- & -- & -- \\\hline
5 & \cite{sunil2023tomato} & 25.60 M & ResNet50 & 99.80 & -- & -- & -- \\\hline
6 & \cite{karimanzira2025context} & 5.70 M & ViT & 96.50 & 93.90 & 96.70 & 94.20 \\\hline
7 & \textbf{Our} & \textbf{2.47 M} & \textbf{CoAtNet-DeepMoE} & \textbf{99.83} & \textbf{99.85} & \textbf{99.76} & \textbf{99.80} \\\hline
\end{tabular}}
\end{table*}

As observed, CoAtNet-DeepMoE again achieves the best performance across all metrics. While Swin Transformer \cite{chen2024using} and ResNet50 \cite{sunil2023tomato} report strong accuracies of 99.1\% and 99.8\%, respectively, they still fall short of the 99.83\% achieved by CoAtNet-DeepMoE. Additionally, their parameter sizes are approximately 11 and 10 times larger than ours, respectively. Furthermore, although the ViT model in \cite{karimanzira2025context} is lightweight with 5.70M parameters, it yields noticeably lower Accuracy, Precision, Recall, and F1-score, falling behind by 3.3\%, 6.0\%, 3.1\%, and 5.6\%, respectively. In contrast, CoAtNet-DeepMoE attains 99.83\% accuracy, 99.85\% precision, 99.76\% recall, and 99.80\% F1-score with only 2.47M parameters.
These results highlight the superior efficiency and strong feature-extraction capability of CoAtNet-DeepMoE, driven by its parameter-efficient mixture-of-experts design. Figure~\ref{fig:3b_Plant_village} further illustrates its advantage in both accuracy and model size compared to other approaches.

\subsection{Ablation Study}
In addition to comparisons with state-of-the-art models, we conducted several ablation studies to validate the robustness and parameter efficiency of the proposed CoAtNet-DeepMoE. To demonstrate robustness, we considered an imbalanced dataset, specifically the PlantVillage dataset, and performed experiments to evaluate the model's ability to generalize and maintain high performance despite class imbalance.

\subsubsection{Cross-Dataset Evaluation}
\begin{figure*}[!htbp]
\centering
\begin{minipage}{0.495\linewidth}
    \centering
    \includegraphics[width=\linewidth]{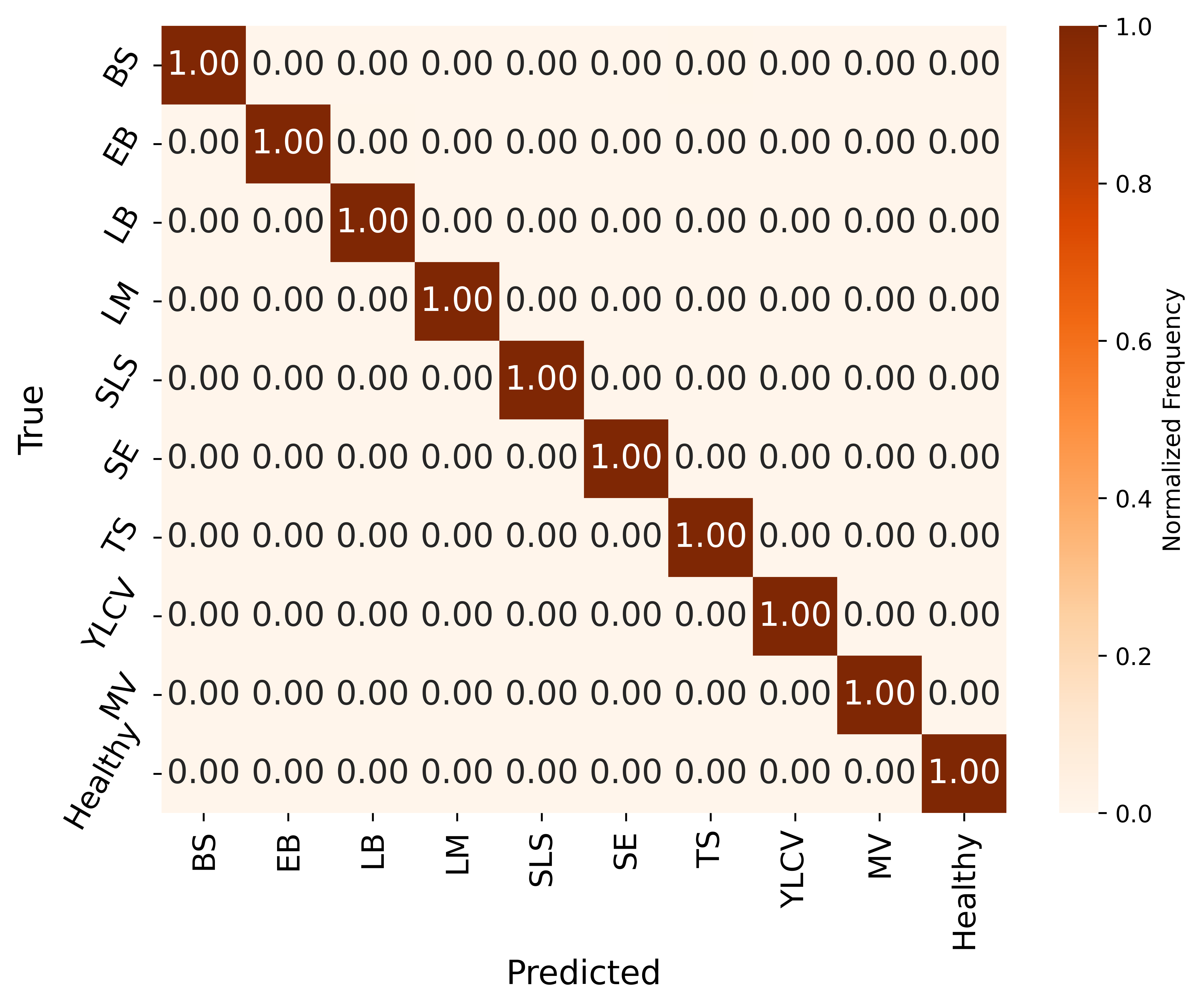}
    \subcaption{Confusion matrix of the Kaggle dataset evaluated on the model trained with the PlantVillage dataset.}
    \label{fig:Kaggle_confusion_matrix_train_on_Plant_village_dataset}
\end{minipage}
\hfill
\begin{minipage}{0.495\linewidth}
    \centering
    \includegraphics[width=\linewidth]{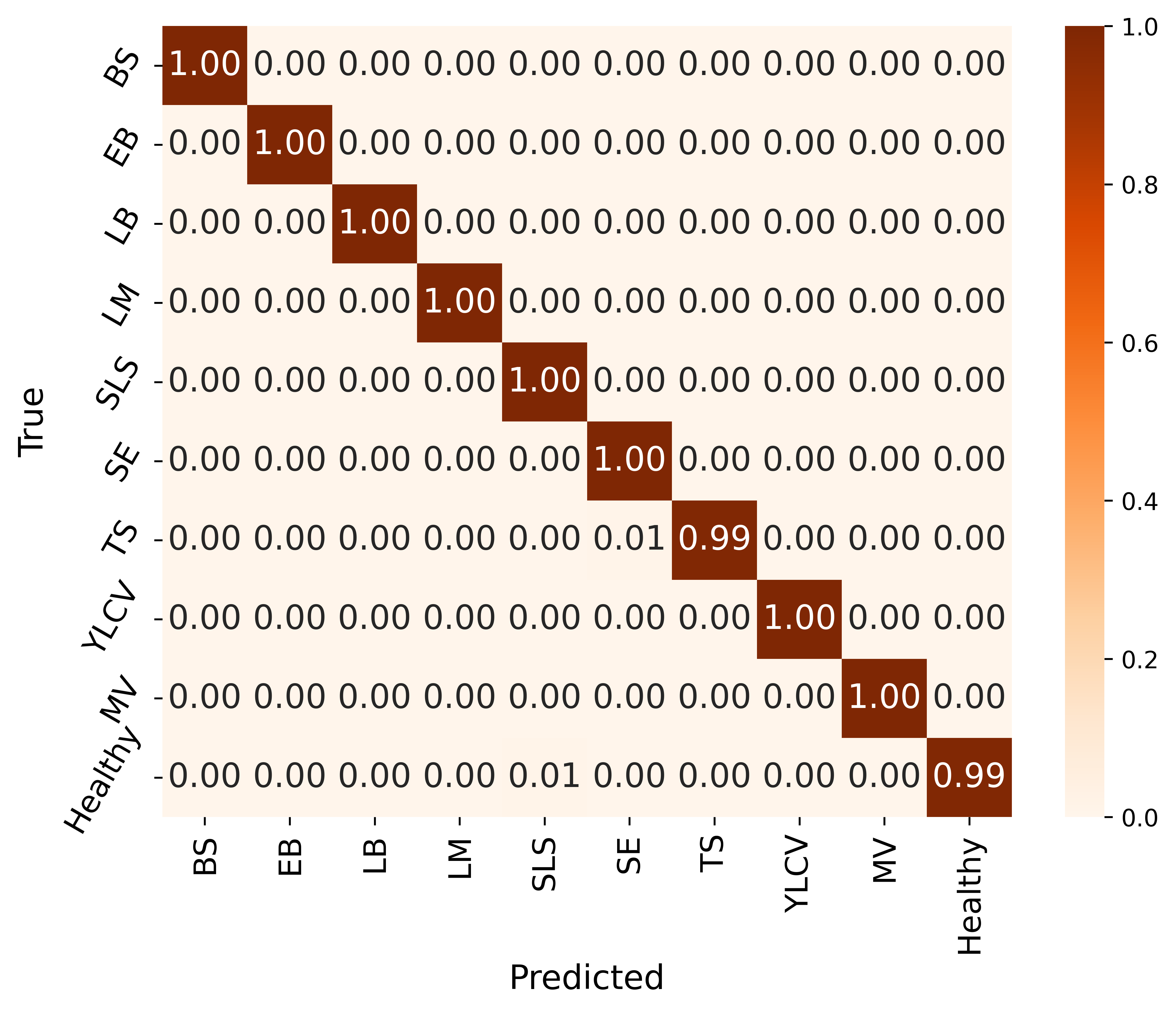}
    \subcaption{Confusion matrix of the PlantVillage dataset evaluated on the model trained with the Kaggle dataset.}
    \label{fig:Plant_village_confusion_matrix_train_on_Kaggle_dataset}
\end{minipage}
\caption{Normalized confusion matrices for cross-dataset evaluation.}
\label{fig:cross_dataset_comparison}
\end{figure*}

To assess the generalization capability of the trained models, we conducted cross-dataset evaluations. Specifically, models trained on the PlantVillage dataset were tested on the Kaggle tomato leaf disease dataset, and conversely, models trained on the Kaggle dataset were tested on the PlantVillage dataset. Both datasets consist of ten classes of tomato leaf diseases. The resulting normalized confusion matrices for these cross-dataset experiments are presented in Figure~\ref{fig:cross_dataset_comparison}. These matrices provide a clear visualization of the model's ability to generalize across datasets with differing image characteristics, including variations in lighting, background, and acquisition conditions. They also indicate which classes are accurately predicted and highlight the classes that are more prone to misclassification under unseen dataset distributions.

\begin{table*}[htbp]
\centering
\small
\setlength{\tabcolsep}{12pt}
\renewcommand{\arraystretch}{1.25}
\caption{Class-wise performance of the proposed model under cross-dataset evaluation. The model is trained on one dataset and evaluated on the other. Accuracy, Precision, Recall, and F1-score are reported as decimal values. The Average row represents the macro-average across all ten classes.}
\label{tab:cross_dataset_performance}
\begin{tabular}{@{}lcccc|cccc@{}}
\toprule
& \multicolumn{4}{c|}{\textbf{PlantVillage $\rightarrow$ Kaggle}}
& \multicolumn{4}{c}{\textbf{Kaggle $\rightarrow$ PlantVillage}} \\ \midrule
\textbf{Class}
& \textbf{Accuracy} & \textbf{Precision} & \textbf{Recall} & \textbf{F1-score} & \textbf{Accuracy} & \textbf{Precision} & \textbf{Recall} & \textbf{F1-score} \\
\midrule
BS       & 1.0000 & 1.0000 & 1.0000 & 1.0000& 0.9994 & 1.0000 & 0.9951 & 0.9975 \\
EB       & 1.0000 & 1.0000 & 1.0000 & 1.0000& 0.9997 & 1.0000 & 0.9953 & 0.9977 \\
LB       & 1.0000 & 1.0000 & 1.0000 & 1.0000& 0.9994 & 0.9949 & 1.0000 & 0.9974 \\
LM       & 1.0000 & 1.0000 & 1.0000 & 1.0000& 1.0000 & 1.0000 & 1.0000 & 1.0000 \\
SLS      & 0.9990 & 0.9901 & 1.0000 & 0.9950& 0.9997 & 1.0000 & 0.9971 & 0.9986 \\
SE       & 0.9990 & 0.9901 & 1.0000 & 0.9950& 1.0000 & 1.0000 & 1.0000 & 1.0000 \\
TS       & 0.9990 & 1.0000 & 0.9900 & 0.9950& 0.9992 & 0.9885 & 1.0000 & 0.9942 \\
YLCV     & 1.0000 & 1.0000 & 1.0000 & 1.0000& 1.0000 & 1.0000 & 1.0000 & 1.0000 \\
MV       & 1.0000 & 1.0000 & 1.0000 & 1.0000& 1.0000 & 1.0000 & 1.0000 & 1.0000 \\
Healthy  & 0.9990 & 1.0000 & 0.9900 & 0.9950& 0.9997 & 1.0000 & 0.9970 & 0.9985 \\
\midrule
\textbf{Average} & \textbf{0.9996} & \textbf{0.9980} & \textbf{0.9980} & \textbf{0.9980}& \textbf{0.9997} & \textbf{0.9983} & \textbf{0.9985} & \textbf{0.9984} \\
\bottomrule
\end{tabular}
\end{table*}

If we observe Table~\ref{tab:cross_dataset_performance}, which presents class-wise accuracy, precision, recall, and F1-score for cross-dataset evaluation, we notice that the BS, SE, and TS classes exhibit slightly lower performance compared to the others, with a few instances misclassified in both datasets. The remaining classes achieve near-perfect scores in one dataset but may have minor errors in the other. Overall, these experiments demonstrate that CoAtNet-DeepMoE maintains high classification performance and effectively generalizes to unseen data, even when trained on imbalanced datasets.

\subsubsection{Inference Cost Analysis}

We evaluated CoAtNet-DeepMoE alongside CoAtNet-Base, ResNet50, and ViT-Tiny under identical hardware and measurement conditions. As shown in Table~\ref{tab:model_cost_analysis}, CoAtNet-DeepMoE requires the fewest parameters (2.47M) and the lowest computational cost (1.34 GFLOPs), being 2.2 to 10.8 times smaller and 1.6 to 6.3 times less compute-intensive than the other models. In terms of raw per-image latency, ViT-Tiny is marginally the fastest on average (3.78 ms vs. 4.04 ms for CoAtNet-DeepMoE); however, CoAtNet-DeepMoE achieves the lowest tail latency, with the best p95 (4.19 ms) and p99 (4.24 ms) values among all four models, indicating more consistent inference behavior across repeated runs. CoAtNet-Base is the slowest and most compute-heavy model overall, at 6.79 ms/img and 8.42 GFLOPs.

\begin{figure}[!htbp]
\centering
\begin{minipage}{0.95\linewidth}
    \centering
    \includegraphics[width=\linewidth]{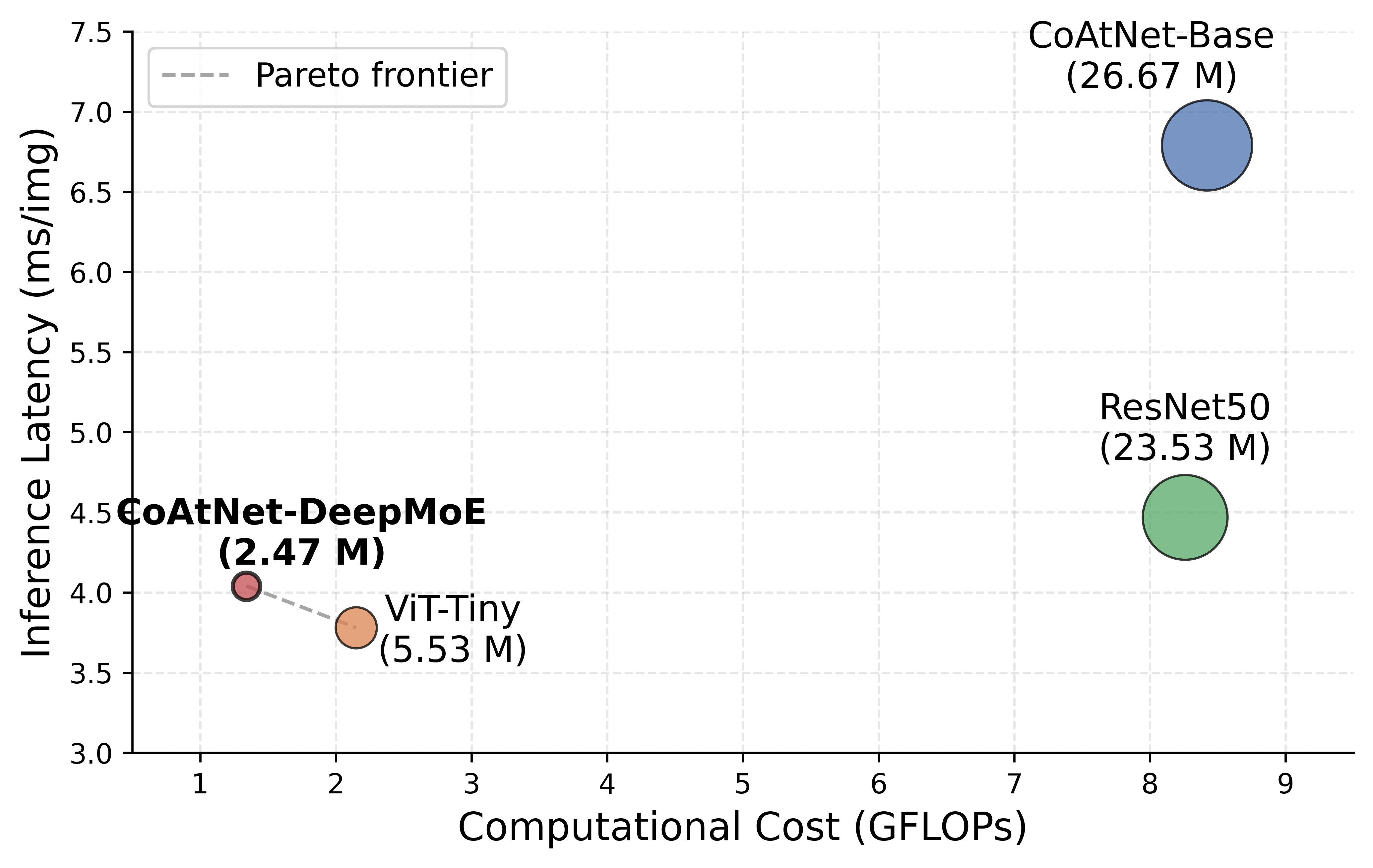}
\end{minipage}
\caption{Cost--latency trade-off among CoAtNet-Base, ResNet50, ViT-Tiny, and the proposed CoAtNet-DeepMoE. The x-axis shows computational cost (GFLOPs), the y-axis shows inference latency (ms/img), and bubble size represents the number of parameters (M). The dashed line marks the Pareto frontier of non-dominated models.}
\label{fig:ablation_comparison}
\end{figure}
 
\begin{table*}[!t]
\centering
\caption{Cost analysis of different models in terms of parameter count (M), computational complexity (GFLOPs), and inference latency (ms/img), measured on the same hardware. p95/p99 denote the 95th/99th-percentile per-image latency over repeated runs. All metrics follow a lower-is-better ($\downarrow$) convention, except FPS, which is higher-is-better ($\uparrow$). Bold values indicate the most efficient performance among the compared models.}
\label{tab:model_cost_analysis}
\begin{tabular}{lcccccc}
\hline
\textbf{Model} & \textbf{Params (M)$\downarrow$} & \textbf{GFLOPs$\downarrow$} & \textbf{Time/img (ms)$\downarrow$} & \textbf{p95 (ms)$\downarrow$} & \textbf{p99 (ms)$\downarrow$} & \textbf{FPS$\uparrow$} \\\hline
CoAtNet-Base & 26.67 & 8.42 & 6.79 & 6.90 & 6.98 & 147.3 \\ \hline
ResNet50     & 23.53 & 8.26 & 4.47 & 4.65 & 4.70 & 223.6 \\ \hline
ViT-Tiny     & 5.53  & 2.15 & \textbf{3.78} & 4.93 & 5.02 & \textbf{264.8} \\ \hline
\textbf{CoAtNet-DeepMoE} & \textbf{2.47} & \textbf{1.34} & 4.04 & \textbf{4.19} & \textbf{4.24} & 247.6 \\ \hline
\end{tabular}
\end{table*}

Figure~\ref{fig:ablation_comparison} visualizes this trade-off, plotting GFLOPs against inference latency with bubble size encoding parameter count. CoAtNet-DeepMoE and ViT-Tiny lie on the Pareto frontier, jointly dominating CoAtNet-Base and ResNet50 in both cost and latency, while CoAtNet-DeepMoE additionally achieves this with less than half the parameters of ViT-Tiny. Overall, CoAtNet-DeepMoE delivers the best parameter and computational efficiency among the compared architectures, while remaining competitive with the fastest model (ViT-Tiny) in latency and throughput, underscoring its suitability for low-resource and real-time applications.

\section{Discussion}
The experimental results across both the Kaggle and PlantVillage tomato disease datasets demonstrate that the proposed CoAtNet-DeepMoE model achieves state-of-the-art performance (Table \ref{tab:performance_comparison_Kaggle} and Table \ref{tab:performance_comparison_Plant_village}) while maintaining exceptionally low computational cost. Compared to traditional convolutional and residual networks (e.g., ResNet50) and recent transformer-based architectures (e.g., ViT-Tiny, CoAtNet-Base), CoAtNet-DeepMoE consistently yields higher accuracy, precision, recall, and F1-score, despite having significantly fewer parameters.

We believe that the DeepSeek-MoE mechanism can have a substantial impact on vision applications for agriculture. In leaf disease classification, for example, certain feature patterns, such as healthy leaf texture or background characteristics, are common across classes and can be efficiently captured by the shared expert space, which acts as a repository for universally relevant information. At the same time, disease-specific variations (e.g., lesion shape, color distortion, or infection boundary) can be processed by the top-$k$ experts selected dynamically by the routing function based on the input. This selective routing enables each expert to specialize in identifying distinct disease-related features while avoiding unnecessary computation. Furthermore, DeepSeek-MoE inherits the core advantages of traditional mixture-of-experts architectures, where a large model is decomposed into several smaller subnetworks (experts). The addition of a shared expert enhances this design, enabling extensive parameter reduction without compromising classification performance. Overall, this structure allows high representational capacity with low computational cost, making it particularly well-suited for agricultural vision tasks.

\textit{This performance gain can be attributed to two key design principles:}

First, the hybrid convolution-attention structure of CoAtNet \cite{dai2021coatnet} provides stronger spatial modeling than pure CNNs and better local feature extraction than pure transformers. Specifically, the CoAtNet model initially applies convolutional layers to extract local features, followed by MbConv blocks for channel-wise feature refinement, SE blocks for enhanced feature recalibration, and attention layers to capture global dependencies. This combination allows the model to simultaneously extract both local and global features, which is critical for accurate disease classification. In contrast, other popular models such as MobileNetV4, EfficientNet-V2-S, and ConvNeXt-V2 rely solely on convolutional operations and do not incorporate attention mechanisms, limiting their ability to model long-range dependencies.

Second, the parameter-efficient DeepSeek-MoE mechanism selectively routes features through specialized experts, enabling richer and more diverse representations while drastically reducing computational overhead compared to traditional mixture-of-experts models. As observed in the ablation study, the compared models require 2.2 to 10.8 times more parameters than CoAtNet-DeepMoE, and in the broader SOTA comparison several models require substantially more parameters yet still achieve lower performance, demonstrating that a higher parameter count does not necessarily translate to better accuracy. This efficient design allows CoAtNet-DeepMoE to maintain high representational capacity with minimal computational cost, making it particularly effective for real-time and resource-constrained applications.

\section{Conclusion}

This study presents CoAtNet-DeepMoE, a parameter-efficient hybrid convolution-attention model with a DeepSeek mixture-of-experts design for tomato leaf disease classification. Evaluated on two benchmark datasets, it achieves SOTA performance while using only 2.47M parameters, the lowest GFLOPs, and the fastest tail latency, outperforming several state-of-the-art models. The confusion matrices (Fig.~\ref{fig:4a_Kaggle_confusion_matrix} and Fig.~\ref{fig:4b_Plant_village_confusion_matrix}) confirm near-perfect classification across all disease categories, and the model demonstrates strong generalizability under different imaging conditions. With its small footprint and low latency, CoAtNet-DeepMoE is well-suited for real-time and edge-based agricultural applications. Future work may focus on field-level deployment, more diverse imagery, and integration with early-warning decision-support systems, further enhancing its practical impact.


\bibliographystyle{unsrtnat}
\bibliography{references}  

\end{document}